\documentclass[conference]{IEEEtran}
\IEEEoverridecommandlockouts
\usepackage{cite}
\usepackage{amsmath,amssymb,amsfonts}
\usepackage{algorithmic}
\usepackage{graphicx}
\usepackage{textcomp}
\usepackage{xcolor}

\usepackage{amsmath}
 \usepackage{multirow}
 \usepackage{booktabs}
 \usepackage{placeins} 
\usepackage{xcolor}
\usepackage{float}
\usepackage{placeins}
\usepackage{booktabs}

\usepackage{longtable}
\usepackage{xspace}

\usepackage{enumitem}
\usepackage{subfigure}
\usepackage{hyperref}

\usepackage{multirow}
\renewcommand\arraystretch{1.3}

\usepackage{url}
\usepackage{comment}

\usepackage{tikz}
\usetikzlibrary{arrows.meta,positioning}

\usepackage{pgfplots}
\usepackage{pgfplotstable}
\usepgfplotslibrary{groupplots}
\pgfplotsset{compat=1.18}

\def\BibTeX{{\rm B\kern-.05em{\sc i\kern-.025em b}\kern-.08em
    T\kern-.1667em\lower.7ex\hbox{E}\kern-.125emX}}
    
\begin{document}

\title{Domain-Adapted Small Language Models for Reliable Clinical Triage
}

\author{\IEEEauthorblockN{Anonymous Authors}}

\author{\IEEEauthorblockN{1\textsuperscript{st} Manar Aljohani}
\IEEEauthorblockA{\textit{Department of Computer Science} \\
\textit{Virginia Tech}\\
Blacksburg, VA, US \\
manara@vt.edu}
\and
\IEEEauthorblockN{2\textsuperscript{nd} Brandon Ho}
\IEEEauthorblockA{\textit{Division of Emergency Medicine} \\
\textit{University of Washington and Seattle Children's Hospital}\\
Seattle, WA, USA \\
Brandon.Ho@seattlechildrens.org}
\and
\IEEEauthorblockN{3\textsuperscript{rd} Kenneth McKinley}
\IEEEauthorblockA{\textit{Children’s National Hospital}\\
Washington DC, USA \\
kmckinley@childrensnational.org}
\and
\IEEEauthorblockN{4\textsuperscript{th} Dennis Ren}
\IEEEauthorblockA{\textit{Children’s National Hospital}\\
Washington DC, USA \\
dmren2@childrensnational.org}
\and
\IEEEauthorblockN{5\textsuperscript{th} Xuan Wang}
\IEEEauthorblockA{\textit{Department of Computer Science} \\
\textit{Virginia Tech}\\
Blacksburg, VA, US \\
xuanw@vt.edu}
}

\maketitle

\begin{abstract}
Accurate and consistent Emergency Severity Index (ESI) assignment remains a persistent challenge in emergency departments, where highly variable free-text triage documentation contributes to mistriage and workflow inefficiencies. This study evaluates whether open-source small language models (SLMs) can serve as reliable, privacy-preserving decision-support tools for clinical triage. We systematically compared multiple SLMs across diverse prompting pipelines and found that clinical vignettes, concise summaries of triage narratives, yielded the most accurate predictions. The SLM, Qwen2.5-7B, demonstrated the strongest balance of accuracy, stability, and computational efficiency. Through large-scale domain adaptation using expert-curated and silver-standard pediatric triage data, fine-tuned Qwen2.5-7B models substantially reduced discordance and clinically significant errors, outperforming all baseline SLMs and advanced proprietary large language models (LLMs, e.g., GPT-4o). These findings highlight the feasibility of institution-specific SLMs for reliable, privacy-preserving ESI decision support and underscore the importance of targeted fine-tuning over more complex inference strategies.
\end{abstract}

\begin{IEEEkeywords}
Emergency Severity Index, emergency department triage, small language models, clinical decision support, QLoRA fine-tuning, privacy-preserving artificial intelligence
\end{IEEEkeywords}

\section{Introduction}
\textit{Triage Accuracy and Variability.} Timely and accurate triage is a cornerstone of emergency department (ED) operations, influencing patient outcomes, resource allocation, and overall efficiency. The Emergency Severity Index (ESI) is a widely used five-level triage system that classifies patients by acuity and expected resource needs, ranging from ESI 1 (immediate life-saving intervention) to ESI 5 (no resources anticipated). ESI 2 includes high-risk patients without immediate resuscitation needs, while ESI 3–5 are primarily distinguished by expected resource use: ESI 3 indicates patients likely to require two or more resources, and ESI 4 those expected to require a single resource \cite{gilboy2020esi}. In practice, ESI assignment is often subjective and variable, especially under high workload or limited information, with studies reporting only $\sim$30–50\% correct triage rates \cite{van2021machine}. For example, one multicenter study found correct assignment in just 34.1\% of pediatric ED visits using ESI~v4 \cite{sax2024emergency}, while another adult ED study reported mistriage in $\sim$32\% of encounters and sensitivity for high-acuity patients of only $\sim$66\% \cite{sax2023evaluation}. An international multi-center assessment reported accuracy rates near $\sim$44–50\% in some settings \cite{mistry2018accuracy}. These misclassifications can delay life-saving care or overburden emergency resources, highlighting the need for reliable data-driven triage support systems.

Recent work on automating ESI prediction has progressed from traditional machine learning models using structured features to deep learning methods incorporating free-text triage notes. Early approaches employed recurrent neural networks such as LSTMs to model clinical narratives \cite{lipton2015learning}, followed by transformer-based encoders such as BERT and ClinicalBERT, which improved performance through large-scale pretraining and domain-specific adaptation \cite{alsentzer2019public}. The emergence of large language models (LLMs) marks the next stage: models like GPT-3 scale to billions of parameters and enable strong zero-shot and few-shot reasoning \cite{brown2020language}, while medical LLMs such as Med-PaLM demonstrate robust clinical understanding without extensive task-specific supervision \cite{singhal2023large}. These advances position LLMs as promising tools for context-aware ESI prediction, particularly when annotated triage data is limited and clinical reasoning is required.

\begin{figure*}[t]
  \centering
    \includegraphics[width=1.0\textwidth]{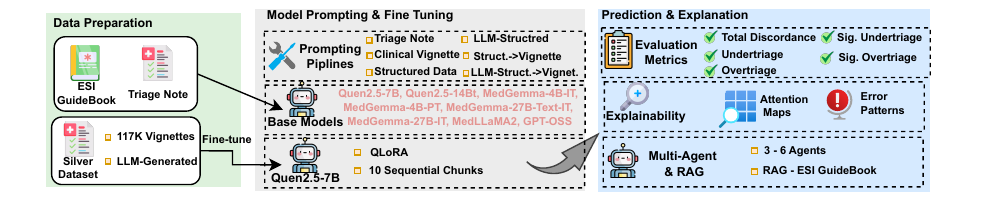}
 \caption{Overview of the methodological pipeline, illustrating data and training sources, silver-standard vignette generation, six prompting pipelines, QLoRA fine-tuning of Qwen2.5-7B, evaluation metrics, explainability analysis, and additional multi-agent and RAG experiments.}
 \label{fig:Method_figure}
\end{figure*}

\textit{Opportunities and Challenges of LLMs for Triage.}
Recent advances in large language models (LLMs) have demonstrated new opportunities to analyze complex unstructured clinical narratives, such as triage notes or patient vignettes \cite{singhal2023large, nori2023capabilities, lu-etal-2024-triageagent}. LLMs show strong abilities in extracting clinical features, reasoning over patient narratives, and generating structured outputs while also raising important concerns about hallucination, robustness, and domain reliability \cite{wu2023med,aljohani-etal-2025-comprehensive}. These strengths suggest potential value in assisting triage by standardizing ESI assignments, reducing cognitive load, and identifying high-risk patients earlier.

However, deploying proprietary LLMs in clinical environments remains challenging. These models are typically closed-box systems, require external API calls that raise HIPAA concerns, and demand substantial compute resources that hinder on-premise deployment \cite{thirunavukarasu2023large, nori2023capabilities}. Critically, their outputs are often inconsistent across repeated runs, which is problematic for safety-critical tasks such as triage that require stability, reproducibility, and transparency \cite{ho2024evaluation}. As a result, these limitations slow real-world operationalization despite strong clinical reasoning capabilities.

\textit{Small Language Models as a Solution.}
Open-source \textit{Small Language Models} (SLMs), typically $\leq$7B parameters, offer a promising alternative for specialized, resource-constrained, and privacy-sensitive healthcare environments \cite{openlm2024tinyllm, dettmers2023qlora}. Recent advances in quantization \cite{dettmers2023qlora}, parameter-efficient fine-tuning (PEFT) \cite{hu2022lora}, and low-rank adaptation methods such as LoRA and QLoRA\cite{dettmers2023qlora} enable efficient domain adaptation while maintaining competitive accuracy. SLMs provide several advantages: they are cost-efficient, interpretable, and adaptable to institution-specific triage data without exposing patient information \cite{dettmers2023qlora,hu2022lora,openlm2024tinyllm}. Their open architectures support reproducibility, transparent evaluation, and community-driven improvement. Because SLMs are lightweight enough for on-premise deployment behind institutional firewalls, they support strict privacy requirements, reduce inference latency, and enable version-controlled reproducibility. Notably, SLMs often produce more stable outputs for rule-oriented clinical decision-support tasks than larger LLMs, especially when fine-tuned on structured or guideline-based workflows. In medical applications, these properties allow SLMs to achieve strong in-domain performance with significantly lower memory, compute, and operational overhead than LLMs, making them well-suited for reliable and privacy-preserving emergency triage systems.

Prior work has explored similar strategies for adapting general-purpose transformers to medical question answering \cite{singhal2023large}, note summarization \cite{jin2023chatgpt}, and radiology report generation \cite{wu2023med}.  However, few studies have examined the fine-tuning of SLMs for structured rule-based decision-support tasks such as clinical ESI triage assignment where safety, privacy, and interpretability are critical. To address this gap, we systematically evaluated multiple open-source SLM frameworks and fine-tuned a 7B-parameter SLM (Qwen2.5-7B) on de-identified retrospective emergency department triage data. We assessed the extent to which a fine-tuned Qwen2.5-7B model can approximate human ESI decision-making and analyze its error patterns (undertriage, overtriage) to identify areas for trust-aware improvement in clinical deployment. Our objective was to assess the feasibility and accuracy of small, transparent language models as a trustworthy alternative for clinical decision support in emergency triage (Figure \ref{fig:Method_figure}).

\section{Materials \& Methods}

\subsection{Data Sources}
The dataset was obtained from de-identified pediatric emergency department records from a large academic children’s hospital. An initial cohort of approximately 117,600 pediatric encounters was screened to create a structured dataset containing key triage variables for each ED visit. Each record includes a de-identified patient index, the Emergency Severity Index (ESI) assigned by clinical staff, patient age in months, and the unstructured free-text triage note describing the presenting complaint. The dataset also includes a hashed patient reference number (FIN) and the documented \textit{Chief Complaint}, summarizing the primary reason for the ED visit as reported by the caregiver or patient and recorded by the triage nurse. Additional fields capture vital signs (e.g., heart rate, temperature, respiratory rate) and physical exam findings. When vital signs were unavailable, a pivot assessment field provided supplemental observations. Finally, each record contains the patient’s past medical history (PMH) as documented during triage.

\subsection{Inclusion/Exclusion Criteria for Our Dataset.}
We included encounters for individuals aged 0–21 years presenting to the pediatric emergency department who underwent standard nursing triage with an assigned ESI level. Encounters were excluded under several conditions. First, cases in which patients bypassed standard triage—such as direct-to-room placement due to high acuity, transport team handoff, or resuscitation bay activation—were removed, as most ESI~1 patients receive immediate physician evaluation without a discretionary ESI determination. Second, encounters with pre-arrival activation (e.g., EMS alerts, trauma or critical care activations, or transfer notifications from referring providers) were excluded because these events predetermine patient placement and add information outside the standard triage workflow. Finally, primary mental or behavioral health presentations were excluded because they follow a distinct triage pathway emphasizing behavioral safety rather than physiologic acuity or resource-based ESI decisions.

\subsection{Data Filtering and Curation}
To ensure data completeness and clinical relevance, we created a filtered subset of encounters ($n = 353$) by retaining records with meaningful clinical information. Encounters were included only if the Chief Complaint field contained substantive clinical content rather than placeholder text (e.g., “see pivot,” “see triage,” or “none”). Valid physiological measurements were required in the Vital Signs field, and the Physical Exam field had to contain documented clinical observations rather than empty entries. The Pivot Assessment field was retained only when it provided clinically relevant supplemental information, typically absent when vital signs were available. Records also required a valid Past Medical History (PMH) entry, excluding those marked “none.” This filtering ensured each encounter contained sufficient structured and unstructured information to support reliable language model prompting and ESI prediction. Requiring complete data allowed us to evaluate model performance under controlled conditions and avoid confounding from heterogeneous missingness. Future work will extend this approach to real-world triage settings by modeling incomplete and evolving data available at presentation.

Importantly, the final evaluation set was not intended to represent the entire ED cohort, but rather a controlled benchmark for comparing model behavior across prompting and fine-tuning strategies under conditions where all input formats could be generated reliably. After applying the initial clinical eligibility criteria, approximately 10,000 encounters contained the core documentation elements needed for vignette generation and model evaluation. The final set of 353 encounters was selected from this complete-documentation subset to support consistent evaluation across raw triage notes, structured fields, and generated clinical vignettes. We acknowledge that this design favors better-documented encounters and may reduce ambiguity relative to routine triage documentation; therefore, results should be interpreted as performance under a controlled evaluation setting rather than as a fully representative estimate of real-world deployment performance.

\subsection{Training Dataset}
The fine-tuning process leveraged two complementary supervision sources. First, the \textbf{ESI Handbook} vignette dataset ($n = 245$) provided expert-curated triage scenarios paired with gold-standard ESI assignments, serving as a high-quality foundation for model training. Second, to address the scarcity of curated vignette–ESI pairs, we generated a large silver-label dataset from de-identified pediatric emergency department records ($n = 117{,}247$) by prompting Qwen2.5-7B to produce concise clinical vignettes from structured triage fields while retaining triage-nurse–assigned ESI labels as ground truth. This silver dataset excludes encounters missing vital signs or physical exam information to ensure clinical completeness. Together, these sources allowed the model to learn both guideline-based reasoning patterns and the variability of real-world clinical presentations, improving robustness and generalization across diverse triage narratives.

\subsection{Evaluation Dataset}
Model evaluation was conducted on a filtered subset of de-identified pediatric emergency department records from the same institution ($n = 353$), which contains high-quality structured and unstructured records. This test set was used consistently across all prompting configurations, including raw triage notes, structured data, and clinical vignettes, to assess model agreement with nurse-assigned ESI labels and to compare undertriage and overtriage patterns across models.

\subsection{Model Selection}
We evaluated multiple open-source and publicly available small and medium-sized language models (SLMs) for Emergency Severity Index (ESI) prediction, including \textbf{Qwen2.5-7B}, \textbf{Qwen2.5-14B-Instruct}, \textbf{MedGemma-4B-IT}, \textbf{MedGemma-4B-PT}, \textbf{MedGemma-27B-Text-IT}, \textbf{MedGemma-27B-IT}, \textbf{MedLLaMA2-7B}, \textbf{GPT-OSS-20B}, and \textbf{GPT-OSS-120B}. All models were accessed through their open-source repositories and deployed locally on secure institutional GPU resources (Table \ref{tab:Models}). 

\begin{table}[t]
\centering
\scriptsize
\setlength{\tabcolsep}{4pt}
\renewcommand{\arraystretch}{1.12}
\caption{Comparison of SLMs on ESI prediction across input variants: total discordance (Disc.), under\-triage (Under), over\-triage (Over), Significant Undertriage (Sig. Under), Significant Overtriage (Sig. Over), and runtime in seconds (Average processing time per patient encounter). Lower is better for Disc./Under/Over/Sig. Under/Sig. Over.}
\label{tab:Models}

\resizebox{\columnwidth}{!}{%
\begin{tabular}{@{}l l r r r r r r@{}}
\toprule
\textbf{Setting} & \textbf{Model} & \textbf{Disc.} & \textbf{Under} & \textbf{Over} & \textbf{Sig. Under} & \textbf{Sig. Over} & \textbf{Time} \\
\midrule

\multirow{9}{*}{Triage note}
& Qwen2.5-7B      & 59.38\% & \textbf{5.97\%} & 53.41\% & \underline{5.40\%} & 11.93\% & \textbf{0.83} \\
& Qwen2.5-14B-Instruct & 60.51\% & \underline{13.64\%} & 46.88\% & 11.36\% & 4.83\% & 12.26 \\
& MedGemma-4B-IT  & 57.67\% & 30.40\% & 26.70\% & 14.77\% & \underline{3.12\%} & 19.73 \\
& MedGemma-4B-PT & 89.50\% & 36.44\% & 53.06\% & 9.04\% & 48.98\% & \underline{11.10} \\
& MedGemma-27B-Text-IT & \underline{53.12\%} & 28.12\% & \underline{25.00\%} & 15.00\% & 23.44\% & 51.32 \\
& MedGemma-27B-IT & \textbf{45.58\%} & 34.19\% & \textbf{11.40\%} & 15.10\% & \textbf{0.57\%} & 46.32 \\
& MedLLaMA2-7B    & 93.35\% & 16.18\% & 77.17\% & \textbf{3.76\%}  & 77.17\% & 15.14 \\
& GPT-OSS-20B  & 80.62\% & 34.38\% & 46.25\% & 8.75\% & 40.62\% & 23.28 \\
& GPT-OSS-120B & 82.65\% & 32.42\% & 50.23\% & 7.76\% & 43.84\% & 19.10 \\
\midrule

\multirow{9}{*}{Clinical vignette}
& Qwen2.5-7B      & 51.70\% & 20.74\% & 30.97\% & 11.36\% & 6.25\%  & \textbf{0.32} \\
& Qwen2.5-14B-Instruct & \underline{45.62\%} & \underline{18.13\%} & 27.49\% & 11.48\% & \underline{4.53\%} & 26.81 \\
& MedGemma-4B-IT  & 67.63\% & 57.55\% & \textbf{10.07\%} & 10.79\% & 8.63\%  & 20.45 \\
& MedGemma-4B-PT & 85.00\% & 22.19\% & 62.81\% & 7.19\% & 52.50\% & 11.74 \\
& MedGemma-27B-Text-IT & 50.00\% & 21.21\% & 28.79\% & 12.12\% & \textbf{4.24\%} & 33.69 \\
& MedGemma-27B-IT & \textbf{45.45\%} & 22.16\% & \underline{23.30\%} & 10.23\% & 5.11\% & 41.56 \\
& MedLLaMA2-7B    & 82.86\% & \textbf{1.71\%}  & 81.14\% & \textbf{1.14\%}  & 77.43\%  & \underline{4.70} \\
& GPT-OSS-20B & 56.89\% & 31.67\% & 25.22\% & 12.90\% & 7.04\% & 9.85 \\
& GPT-OSS-120B & 52.56\% & 22.16\% & 30.40\% & \underline{4.83\%} & 22.73\% & 27.21 \\
\midrule

\multirow{9}{*}{Structured data}
& Qwen2.5-7B      & \underline{61.36\%} & 17.90\% & \underline{43.47\%} & 5.40\%  & \underline{25.00\%}  & \textbf{5.77} \\
& Qwen2.5-14B-Instruct & 62.14\% & \underline{15.61\%} & 46.53\% & 4.91\% & 30.92\% & 27.68 \\
& MedGemma-4B-IT  & 71.94\% & 24.46\% & 47.48\% & \underline{3.24\%}  & 43.88\%  & 21.71 \\
& MedGemma-4B-PT & 89.78\% & 17.34\% & 72.45\% & 5.26\% & 68.11\% & 12.63 \\
& MedGemma-27B-Text-IT & \textbf{55.97\%} & 32.39\% & \textbf{23.58\%} & 13.07\% & \textbf{13.92\%} & 43.90 \\
& MedGemma-27B-IT & 82.41\% & 16.67\% & 65.74\% & 8.33\% & 62.96\% & 13.08 \\
& MedLLaMA2-7B    & 84.29\% & \textbf{1.71\%}  & 82.57\% & \textbf{0.29\%}  & 82.29\%  & \underline{9.45} \\
& GPT-OSS-20B & 85.01\% & 17.00\% & 68.01\% & 5.19\% & 63.11\% & 18.15 \\
& GPT-OSS-120B & 76.70\% & 18.18\% & 58.52\% & 5.11\% & 50.00\% & 61.60 \\
\midrule

\multirow{9}{*}{LLM-structured}
& Qwen2.5-7B      & \underline{56.69\%} & \textbf{8.72\%} & 47.97\% & 7.27\%  & \underline{13.37\%}  & \underline{6.35} \\
& Qwen2.5-14B-Instruct & 58.67\% & 22.25\% & 36.42\% & 6.65\% & 24.57\% & 28.28 \\
& MedGemma-4B-IT  & 68.47\% & 22.44\% & 46.02\% & \underline{2.56\%}  & 40.91\%  & 21.46 \\
& MedGemma-4B-PT & 86.02\% & 17.70\% & 68.32\% & 4.04\% & 62.42\% & 12.27 \\
& MedGemma-27B-Text-IT & \textbf{55.27\%} & 42.17\% & \textbf{13.11\%} & 14.81\% & \textbf{0.57\%} & 54.48 \\
& MedGemma-27B-IT & 64.52\% & 34.41\% & \underline{30.11\%} & 10.75\% & 25.81\% & 9.61 \\
& MedLLaMA2-7B    & 86.99\% & \underline{12.43\%} & 74.57\% & \textbf{2.02\%}  & 73.41\%  & \textbf{4.10} \\
& GPT-OSS-20B & 79.23\% & 15.77\% & 63.46\% & 3.46\% & 55.00\% & 16.51 \\
& GPT-OSS-120B & 79.25\% & 13.54\% & 65.71\% & 4.61\% & 61.10\% & 58.40 \\
\midrule

\multirow{9}{*}{Vignette $\leftarrow$ Struct.}
& Qwen2.5-7B      & 52.56\% & 21.02\% & \underline{31.53\%} & 8.52\%  & \textbf{9.09\%}  & \textbf{0.32} \\
& Qwen2.5-14B-Instruct & \underline{50.29\%} & \underline{14.12\%} & 36.18\% & 9.12\% & \underline{9.41\%} & 22.54 \\
& MedGemma-4B-IT  & 62.39\% & 28.21\% & 34.19\% & 9.40\%  & 29.91\%  & 18.91 \\
& MedGemma-4B-PT & 86.49\% & 23.31\% & 63.18\% & 6.42\% & 55.41\% & 12.02 \\
& MedGemma-27B-Text-IT & 62.08\% & 22.94\% & 39.14\% & 9.79\% & 11.93\% & 25.32 \\
& MedGemma-27B-IT & \textbf{48.15\%} & 19.44\% & \textbf{28.70\%} & 11.11\% & 13.89\% & 12.60 \\
& MedLLaMA2-7B    & 75.07\% & \textbf{6.02\%}  & 69.05\% & \textbf{4.58\%}  & 44.13\%  & \underline{6.21} \\
& GPT-OSS-20B & 65.79\% & 23.39\% & 42.40\% & 13.16\% & 12.28\% & 11.91 \\
& GPT-OSS-120B & 63.07\% & 19.03\% & 44.03\% & \underline{5.11\%} & 33.52\% & 41.52 \\
\midrule

\multirow{9}{*}{Vignette $\leftarrow$ LLM-Struct.}
& Qwen2.5-7B      & 56.25\% & \underline{8.52\%} & 47.73\% & \underline{5.11\%}  & 21.31\%  & \textbf{0.33} \\
& Qwen2.5-14B-Instruct & \underline{54.76\%} & 17.87\% & 36.89\% & 8.93\% & \textbf{10.09\%} & 27.58 \\
& MedGemma-4B-IT  & 63.27\% & 32.65\% & \textbf{30.61\%} & 7.14\%  & 26.02\%  & 18.54 \\
& MedGemma-4B-PT & 85.80\% & 21.77\% & 64.04\% & 5.36\% & 57.73\% & 12.24 \\
& MedGemma-27B-Text-IT & 59.36\% & 27.19\% & \underline{32.16\%} & 11.11\% & 14.33\% & 25.58 \\
& MedGemma-27B-IT & \textbf{53.56\%} & 19.94\% & 33.62\% & 9.12\% & \underline{10.83\%} & 39.52 \\
& MedLLaMA2-7B    & 81.45\% & \textbf{6.09\%}  & 75.36\% & \textbf{3.48\%}  & 62.03\%  & \underline{8.85} \\
& GPT-OSS-20B & 64.52\% & 20.79\% & 43.73\% & 10.75\% & 13.62\% & 10.09 \\
& GPT-OSS-120B & 69.31\% & 34.16\% & 35.15\% & 8.91\% & 29.21\% & 10.89 \\
\bottomrule
\end{tabular}%
}

\begin{minipage}{\columnwidth}
\vspace{0.3em}
\textit{Abbreviations:} “LLM-structured” = fields extracted by an LLM; “Vignette $\leftarrow$ Struct.” = vignette generated from structured data; “Vignette $\leftarrow$ LLM-Struct.” = vignette generated from LLM-structured data.
\end{minipage}
\vspace{-15pt}
\end{table}

\subsection{Prompting Framework}
We evaluated multiple prompting pipelines to assess how input format and intermediate representations affect ESI prediction accuracy. The experiments were organized into the following pipelines:

\begin{itemize}[topsep=1pt, itemsep=1pt, leftmargin=10pt]
    \item \textbf{Raw Triage Record $\rightarrow$ ESI Prediction:} Direct prompting with unstructured triage notes to predict ESI levels.
    \item \textbf{Raw Triage Record $\rightarrow$ Clinical Vignette $\rightarrow$ ESI Prediction:} Models first summarized triage notes into concise clinical vignettes, then predicted ESI.
    \item \textbf{Human-Structured Data $\rightarrow$ ESI Prediction:} Direct prompting with human-provided structured fields to generate ESI levels.
    \item \textbf{Raw Triage Record $\rightarrow$ Structured Data $\rightarrow$ ESI Prediction:} Models extracted structured fields (Chief Complaint, Vital Signs, Physical Exam) and predicted ESI from these structured outputs.
    \item \textbf{Human-Structured Data $\rightarrow$ Clinical Vignette $\rightarrow$ ESI Prediction:} Structured data were used to generate clinical vignettes before ESI prediction.
    \item \textbf{LLM-Generated Structured Data $\rightarrow$ Clinical Vignette $\rightarrow$ ESI Prediction:} Clinical vignettes were generated from LLM-extracted structured data and used for ESI prediction.
\end{itemize}

This design isolates the effects of (1) unstructured vs.\ structured inputs, (2) human- vs.\ LLM-generated intermediate representations, and (3) summarization fidelity on downstream ESI prediction accuracy.

\subsection{Structured Data and Clinical Vignette Generation}
For structured data extraction, models identified and organized relevant information into predefined fields. For vignette generation, models produced concise summaries emphasizing clinical context and patient acuity. To assess vignette quality, a pediatric emergency physician reviewed 50 randomly selected vignette--source note pairs for clinical faithfulness to the original triage documentation. The physician judged 90\% of vignettes to be clinically faithful representations of the source notes. Among the 10\% with discrepancies, most reflected omission of information present in the original note rather than inaccuracies or distortions of the clinical history. No instances of hallucinated clinical information were identified. These findings suggest that the vignette generation process preserved the core clinical content of the original triage documentation while introducing minimal risk of fabricated information.

\subsection{Fine-Tuning of Qwen2.5-7B}
To improve clinical alignment and ESI prediction accuracy, we fine-tuned the \textbf{Qwen2.5-7B} model using expert-curated data and automatically generated (“silver”) clinical vignettes. The silver dataset was \textbf{partitioned into 10 equal chunks} ($\approx$11.7k examples each), and fine-tuning proceeded sequentially: the model was first trained on Chunk~1 and then continued from each checkpoint through Chunk~10. This approach enabled manageable GPU memory usage, stable training across diverse vignettes, and monitoring of model behavior over time.

Fine-tuning used \textit{Quantized Low-Rank Adaptation (QLoRA)} \cite{dettmers2023qlora} to support memory-efficient multi-GPU training. Expert and silver samples were combined through proportional sampling to balance guideline-based supervision with broader clinical variability. Training ran for multiple epochs with early stopping based on validation loss. A consistent preprocessing pipeline and fixed hyperparameters were applied across all chunks, with checkpoints and evaluation metrics logged after each stage to monitor convergence and detect performance drift. To improve reproducibility, Table~\ref{tab:hyperparameters} reports the main hyperparameters used for QLoRA-based fine-tuning. The same preprocessing and training configuration was applied across sequential fine-tuning chunks, with an 85/15 stratified train-validation split within each chunk and model selection based on validation loss.

\begin{table}[t]
\centering
\scriptsize
\setlength{\tabcolsep}{3.5pt}
\caption{Hyperparameters used for QLoRA fine-tuning of Qwen2.5-7B models for ESI prediction from clinical vignettes.}
\label{tab:hyperparameters}
\begin{tabular}{@{}p{0.43\linewidth}p{0.50\linewidth}@{}}
\toprule
\textbf{Hyperparameter} & \textbf{Value} \\
\midrule
Base model & Qwen/Qwen2.5-7B \\
Fine-tuning method & QLoRA \\
Quantization & 4-bit NF4 with double quantization \\
Compute precision & bfloat16 \\
LoRA rank ($r$) & 16 \\
LoRA alpha & 16 \\
LoRA dropout & 0.15 \\
LoRA bias & None \\
Target modules & q\_proj, k\_proj, v\_proj, o\_proj, gate\_proj, up\_proj, down\_proj \\
Task type & Causal language modeling \\
Maximum sequence length & 1024 tokens \\
Train/validation split & 85\% / 15\%, stratified by ESI label \\
Random seed & 42 \\
Per-device batch size & 1 \\
Gradient accumulation steps & 8 \\
Effective batch size & 8 sequences per optimizer update \\
Maximum epochs per chunk & 50 \\
Early stopping & Validation loss; patience = 3 evaluations \\
Learning rate & $1 \times 10^{-5}$ \\
Learning-rate scheduler & Cosine \\
Warmup ratio & 0.03 \\
Optimizer & paged AdamW 8-bit \\
Weight decay & 0.10 \\
Label smoothing & 0.05 \\
Evaluation frequency & Every 100 steps \\
Checkpoint save frequency & Every 100 steps \\
Best-model criterion & Lowest validation loss \\
Gradient checkpointing & Enabled \\
\bottomrule
\end{tabular}
\end{table}

\subsection{Evaluation Metrics}
Model performance was evaluated as agreement with nurse-assigned ESI levels, which reflect the operational triage standard used in routine emergency department care. Nurse-assigned ESI is not a perfect gold standard and may contain variability across clinicians and cases; therefore, our objective was not to determine definitive clinical correctness, but to assess whether SLMs can reproduce routine clinical triage decisions and identify patterns of disagreement, undertriage, and overtriage. Performance was measured using total discordance, undertriage, overtriage, and significant under- and overtriage rates, with \textbf{lower values indicating better performance}. Because ESI is an ordinal scale in which lower numeric values indicate higher acuity (ESI 1 = highest acuity; ESI 5 = lowest acuity), a prediction with a larger ESI number than the reference label represents lower predicted acuity and is counted as undertriage, whereas a prediction with a smaller ESI number represents higher predicted acuity and is counted as overtriage.

\textbf{Total discordance} quantifies the overall error rate by measuring the proportion of misclassified samples among all evaluated texts. It is computed as:
\begin{equation}
\text{Total Discordance} = \frac{\text{Total Misclassifications}}{\text{Total Number of Texts}}
\end{equation}
where \textit{Total Misclassifications} is the number of incorrectly classified queries and \textit{Total Number of Texts} is the total number analyzed.

The \textbf{Undertriage rate} refers to cases where the model predicts a lower acuity than the reference label (predicted label $>$ reference label):
\begin{equation}
\text{Undertriage} = \frac{\text{Number of Predictions} > \text{reference Labels}}{\text{Total Number of Texts}}
\end{equation}

Conversely, the \textbf{Overtriage rate} measures how often the model assigns a higher acuity than warranted (predicted label $<$ reference label):
\begin{equation}
\text{Overtriage} = \frac{\text{Number of Predictions} < \text{Reference Labels}}{\text{Total Number of Texts}}
\end{equation}

The \textbf{Significant Undertriage rate} captures critical misclassifications—cases where the reference label is ESI 2 but predicted as ESI 3, 4, or 5:
\begin{equation}
\begin{aligned}
\text{Significant Undertriage} &= \frac{ \substack{\text{Predicted as 3, 4, or 5}\\ \text{when reference label is 2}} }{ \text{Total number of texts} }
\end{aligned}
\end{equation}

Finally, the \textbf{Significant Overtriage rate} identifies severe overestimations where less critical cases (reference label = 3, 4, or 5) are predicted as ESI 1 or 2:
\begin{equation} \begin{aligned} \text{Significant Overtriage} &= \frac{ \substack{\text{Predicted as 1 or 2}\\ \text{when reference label is 3, 4, or 5}} }{ \text{Total number of texts} } \end{aligned} \end{equation}

\subsection{Explainability}
To better understand cues influencing model predictions, we conducted an exploratory analysis of token-level attention patterns produced by the fine-tuned Qwen2.5-7B model. Because Qwen models are open-source, their internal attention distributions can be examined, enabling qualitative inspection of which parts of the triage narrative the model emphasizes during prediction. Attention visualization was used to identify textual signals associated with model decisions. Although attention weights do not provide causal explanations of model reasoning, they can offer insight into patterns the model prioritizes. We focused on encounters labeled ESI-2 and ESI-3, one of the challenging triage distinctions, and examined token-level attention across correctly and incorrectly classified encounters.

\section{Results}
\subsection{Base Model Comparison}
Our analysis revealed consistent trends across input formats and model families. \textbf{Raw triage data} produced the weakest performance due to noise and irrelevant content, while \textbf{clinical vignettes yielded the most accurate ESI predictions} by preserving essential clinical context in a compact format. Using \textbf{structured data alone} reduced accuracy, likely because narrative cues were lost, and \textbf{model-generated structured data} performed worse than human-curated fields, suggesting distortions or omissions during extraction. Across all settings, \textbf{overtriage} remained the most common error pattern but was reduced when clinical vignettes were used.

\textbf{Across all six input settings, Qwen2.5-7B achieved the best balance of accuracy, interpretability, and efficiency}. The model consistently produced clean structured outputs—especially for vital signs and physical exam fields—with minimal hallucinations. In contrast, \textbf{MedGemma} and \textbf{MedLLaMA2} variants often introduced extraneous or clinically implausible findings and showed higher overtriage rates. Larger models such as \textbf{GPT-OSS-20B} and \textbf{GPT-OSS-120B} offered no meaningful accuracy gains despite substantially higher computational cost.

Table~\ref{tab:Models} summarizes results across models and input variants. \textbf{Qwen2.5-7B offers the strongest foundation for institution-specific fine-tuning, achieving favorable trade-offs across total discordance, undertriage, overtriage, and clinically significant error rates}. Performance is particularly strong in clinically realistic settings—especially vignette-based inputs—where discordance remains around 50–52\% with very low significant undertriage.

Efficiency is a key advantage: \textbf{Qwen2.5-7B is the fastest model evaluated}, with inference typically under one second per encounter—more than an order of magnitude faster than instruction-tuned variants and 20–150$\times$ faster than larger models such as MedGemma-27B or GPT-OSS-120B. Although some larger models occasionally achieve lower errors in isolated settings, these gains are inconsistent and accompanied by substantial latency and higher rates of significant misclassification, limiting their suitability for real-time ED deployment. The key consideration for inference time is whether responses can be returned within the same triage interaction without disrupting clinical workflow.

Overall, the consistency, safety-aligned error patterns, and computational efficiency of \textbf{Qwen2.5-7B} make it the most practical and clinically viable SLM for downstream fine-tuning on pediatric emergency triage data.

\subsection{Fine-Tuning of Qwen2.5 Models}
\begin{figure*}[h]
\centering
\scriptsize
\begin{tikzpicture}
\begin{groupplot}[
    group style={
        group size=4 by 2,
        horizontal sep=0.8cm,   
        vertical sep=1.0cm,     
    },
    height=4.0cm,
    width=0.30\textwidth,       
    xlabel={},
    xticklabel style={rotate=45,anchor=east,font=\scriptsize},
    symbolic x coords={GPT-4o, Base, ESI,2k,5k,10k,C1,C2,C3,C4,C5,C6,C7,C8,C9,C10},
    xtick=data,
    ymin=0,
    grid=both,
    grid style={dashed,gray!20},
    title style={font=\bfseries\footnotesize},
]


\nextgroupplot[
    title={A. 7B Discordance (\%)},
    ylabel={\%},
    ymin=0, ymax=60,
    extra x ticks={GPT-4o},
    extra x tick labels={}
]

\addplot+[only marks, mark=*,thick] coordinates {
    (Base,51.70) (ESI,50.00) (2k,42.33) (5k,38.92) (10k,35.51)
    (C1,31.53) (C2,30.40) (C3,26.99) (C4,27.56)
    (C5,27.56) (C6,25.85) (C7,25.85)
    (C8,26.70) (C9,26.70) (C10,26.70)
    
};
\draw[green!40!black, thick]
    (axis cs:GPT-4o,56.98) --
    (axis cs:C10,56.98);
\node[green!40!black, font=\scriptsize, anchor=north]
    at (axis cs:C3,56.98) {GPT-4o};

\nextgroupplot[
    title={B. 7B Undertriage \& Overtriage},
    ylabel={\%},
    legend style={font=\scriptsize,at={(0.5,1.32)},anchor=south,legend columns=2},
    ymin=0, ymax=60
]

\addplot+[only marks, mark=triangle*,thick] coordinates {
    (Base,20.74) (ESI,13.35) (2k,21.59) (5k,21.59) (10k,17.05)
    (C1,18.47) (C2,18.75) (C3,17.90)
    (C4,17.90) (C5,13.92) (C6,13.35)
    (C7,13.35) (C8,11.65) (C9,11.65)
    (C10,11.65) 
};
\addlegendentry{Undertriage}
\draw[green!40!black, thick]
    (axis cs:GPT-4o,7.41) --
    (axis cs:C10,7.41);
\node[green!40!black, font=\scriptsize, anchor=north]
    at (axis cs:C3,7.41) {GPT-4o Undertriage};

\addplot+[only marks, mark=square*,thick] coordinates {
    (Base,30.97) (ESI,36.65) (2k,20.74) (5k,17.33) (10k,18.47)
    (C1,13.07) (C2,11.65) (C3,9.09)
    (C4,9.66) (C5,13.64) (C6,12.50)
    (C7,12.50) (C8,15.06) (C9,15.06)
    (C10,15.06) 
};
\addlegendentry{Overtriage}

\draw[green!40!black, thick]
    (axis cs:GPT-4o,49.57) --
    (axis cs:C10,49.57);
\node[green!40!black, font=\scriptsize, anchor=north]
    at (axis cs:C3,49.57) {GPT-4o Overtriage};


\nextgroupplot[
    title={C. 7B Significant Errors (\%)},
    ylabel={\%},
    legend style={font=\scriptsize,at={(0.5,1.32)},anchor=south,legend columns=2},
]

\addplot+[only marks, mark=diamond*,thick] coordinates {
    (Base,11.36) (ESI,2.84) (2k,11.36) (5k,12.22) (10k,8.52)
    (C1,8.52) (C2,8.52) (C3,7.10)
    (C4,7.95) (C5,5.11) (C6,6.25)
    (C7,6.25) (C8,4.26) (C9,4.26)
    (C10,4.26) 
};
\addlegendentry{Sig.\ Under}
\draw[green!40!black, thick]
    (axis cs:GPT-4o,2.28) --
    (axis cs:C10,2.28);
\node[green!40!black, font=\scriptsize, anchor=south]
    at (axis cs:C3,2.28) {GPT-4o Sig.\ Under};

\addplot+[only marks, mark=*,thick] coordinates {
    (Base,6.25) (ESI,31.53) (2k,7.10) (5k,4.83) (10k,6.82)
    (C1,3.41) (C2,1.99) (C3,2.56)
    (C4,1.99) (C5,5.11) (C6,2.56)
    (C7,2.56) (C8,5.40) (C9,5.40)
    (C10,5.40) 
};
\addlegendentry{Sig.\ Over}
\draw[green!40!black, thick]
    (axis cs:GPT-4o,27.64) --
    (axis cs:C10,27.64);
\node[green!40!black, font=\scriptsize, anchor=north]
    at (axis cs:C3,27.64) {GPT-4o Sig.\ Over};


\nextgroupplot[
    title={D. 7B Inference Time (s/enc)},
    ylabel={Seconds},
    ymin=0, ymax=7
]

\addplot+[only marks, mark=o,thick] coordinates {
    (Base,0.32) (ESI,0.28) (2k,0.15) (5k,0.15) (10k,0.16)
    (C1,0.25) (C2,0.15) (C3,0.15)
    (C4,0.15) (C5,0.24) (C6,0.15)
    (C7,0.16) (C8,0.15) (C9,0.15)
    (C10,0.15) 
};

\draw[green!40!black, thick]
    (axis cs:GPT-4o,6.15) --
    (axis cs:C10,6.15);
\node[green!40!black, font=\scriptsize, anchor=north]
    at (axis cs:C3,6.15) {GPT-4o};

\nextgroupplot[
    title={E. 14B Discordance (\%)},
    ylabel={\%},
    symbolic x coords={Base,C1,C2,C3,C4,C5,C6,C7,C8,C9,C10},
    xtick=data,
    ymin=0, ymax=60
]

\addplot+[only marks, mark=*,thick] coordinates {
    (Base,45.62) (C1,31.12) (C2,30.55) (C3,26.51)
    (C4,26.80) (C5,25.94) (C6,27.38)
    (C7,29.11) (C8,32.85) (C9,27.38)
    (C10,27.38) 
};

\draw[green!40!black, thick]
    (axis cs:GPT-4o,56.98) --
    (axis cs:C10,56.98);
\node[green!40!black, font=\scriptsize, anchor=north]
    at (axis cs:C3,56.98) {GPT-4o};

\nextgroupplot[
    title={F. 14B Undertriage \& Overtriage},
    ylabel={\%},
    symbolic x coords={Base,C1,C2,C3,C4,C5,C6,C7,C8,C9,C10},
    xtick=data,
    legend style={font=\scriptsize,at={(0.5,1.32)},anchor=south,legend columns=2},
    ymin=0, ymax=60
]

\addplot+[only marks, mark=triangle*,thick] coordinates {
    (Base,18.13) (C1,14.70) (C2,15.56) (C3,17.58)
    (C4,15.85) (C5,14.70) (C6,14.70)
    (C7,13.26) (C8,12.68) (C9,14.70)
    (C10,14.70) 
};

\draw[green!40!black, thick]
    (axis cs:GPT-4o,7.41) --
    (axis cs:C10,7.41);
\node[green!40!black, font=\scriptsize, anchor=north]
    at (axis cs:C3,7.41) {GPT-4o Undertriage};

\addplot+[only marks, mark=square*,thick] coordinates {
    (Base,27.49) (C1,16.43) (C2,14.99) (C3,8.93)
    (C4,10.95) (C5,11.24) (C6,12.68)
    (C7,15.85) (C8,20.17) (C9,12.68)
    (C10,12.68) 
};

\draw[green!40!black, thick]
    (axis cs:GPT-4o,49.57) --
    (axis cs:C10,49.57);
\node[green!40!black, font=\scriptsize, anchor=north]
    at (axis cs:C3,49.57) {GPT-4o Overtriage};


\nextgroupplot[
    title={G. 14B Significant Errors (\%)},
    ylabel={\%},
    symbolic x coords={Base,C1,C2,C3,C4,C5,C6,C7,C8,C9,C10},
    xtick=data,
    legend style={font=\scriptsize,at={(0.5,-0.38)},anchor=north,legend columns=2},
    ymin=0, ymax=30
]

\addplot+[only marks, mark=diamond*,thick] coordinates {
    (Base,11.48) (C1,8.07) (C2,8.07) (C3,7.49)
    (C4,7.78) (C5,7.49) (C6,7.78)
    (C7,4.90) (C8,5.76) (C9,7.78)
    (C10,7.78) 
};
\draw[green!40!black, thick]
    (axis cs:GPT-4o,2.28) --
    (axis cs:C10,2.28);
\node[green!40!black, font=\scriptsize, anchor=south]
    at (axis cs:C3,2.28) {GPT-4o Sig.\ Under};

\addplot+[only marks, mark=*,thick] coordinates {
    (Base,4.53) (C1,3.75) (C2,1.15) (C3,1.44)
    (C4,1.44) (C5,3.17) (C6,2.02)
    (C7,4.03) (C8,5.76) (C9,2.02)
    (C10,2.02) 
};
\draw[green!40!black, thick]
    (axis cs:GPT-4o,27.64) --
    (axis cs:C10,27.64);
\node[green!40!black, font=\scriptsize, anchor=north]
    at (axis cs:C3,27.64) {GPT-4o Sig.\ Over};


\nextgroupplot[
    title={H. 14B Inference Time (s/enc)},
    ylabel={Seconds},
    symbolic x coords={Base,C1,C2,C3,C4,C5,C6,C7,C8,C9,C10},
    xtick=data,
]

\addplot+[only marks, mark=o,thick] coordinates {
    (Base,26.81) (C1,0.36) (C2,0.69) (C3,0.41)
    (C4,0.41) (C5,0.83) (C6,0.40)
    (C7,0.50) (C8,0.42) (C9,0.42)
    (C10,0.41) 
};
\draw[green!40!black, thick]
    (axis cs:GPT-4o,6.15) --
    (axis cs:C10,6.15);
\node[green!40!black, font=\scriptsize, anchor=north]
    at (axis cs:C3,6.15) {GPT-4o};
\end{groupplot}
\end{tikzpicture}
\caption{Fine-tuning comparisons for Qwen2.5 models on ESI prediction from clinical vignettes. Top row: Qwen2.5-7B; bottom row: Qwen2.5-14B-Instruct. X-axis settings are discrete: \textbf{Base} = unfine-tuned; \textbf{ESI} = fine-tuned on synthetic ESI Handbook vignettes; \textbf{2k}, \textbf{5k}, \textbf{10k} = fine-tuned on 2,000, 5,000, and 10,000 silver-label examples; \textbf{C1--C10} = sequential chunk-based fine-tuning stages. Panels show discordance, under-/over-triage, significant errors, and inference time per encounter. Significant errors differ from the reference ESI by $\geq$2 acuity levels. Green horizontal lines mark GPT-4o on the same test set. Lower is better for all metrics.}
\label{fig:qwen_finetune_lineplots}
\end{figure*}
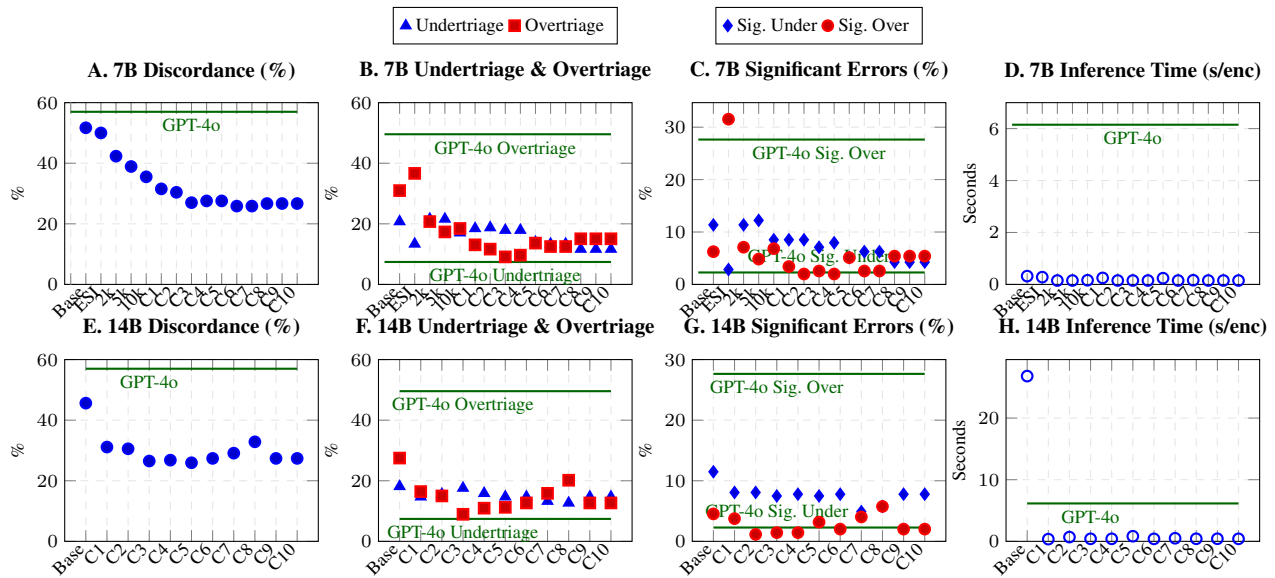

\textbf{Fine-tuning substantially improved Qwen2.5-7B’s ESI prediction from clinical vignettes}, yielding more stable reasoning, fewer hallucinated findings, and better discrimination between high- and low-acuity cases (Figure~\ref{fig:qwen_finetune_lineplots}). Domain-specific fine-tuning on the institutional silver-label dataset produced the largest gains: discordance decreased from 42.33\% (2k examples) to 35.51\% (10k) and reached \textbf{25.85\%} in chunk-level models (Chunks~6–7), with low undertriage (13.35\%), overtriage (12.50\%), and significant overtriage (2.56\%), indicating effective learning of institution-specific triage patterns.  


\FloatBarrier

Fine-tuning on ESI Handbook synthetic vignettes produced more modest benefits: although undertriage decreased, overtriage remained high (36.65\%). In contrast, the real-world silver-label data derived from de-identified pediatric emergency department records—particularly in chunked training—yielded stable, monotonic improvements across all error categories. For comparison, \textbf{Qwen2.5-14B-Instruct} fine-tuned models achieved competitive accuracy (best: 25.94\% discordance), but at substantially higher inference cost (0.36–0.83\,s/encounter). Qwen2.5-7B retained \textbf{exceptionally fast inference} (0.15–0.28\,s/encounter), reinforcing its suitability for real-time emergency department workflows.

These results show that \textbf{Qwen2.5-7B is the most effective and practical model for institution-specific fine-tuning}, offering the strongest combination of accuracy, safety-aligned error behavior, inference speed, and training stability.

\subsection{Explainability of the Fine-tuned Qwen2.5 Models}
\begin{figure}[h]
\centering
\begin{minipage}{0.24\textwidth}
\centering
\includegraphics[width=\linewidth]{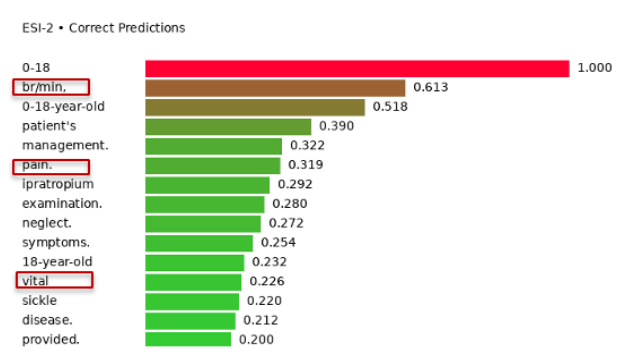}
\end{minipage}
\begin{minipage}{0.24\textwidth}
\centering
\includegraphics[width=\linewidth]{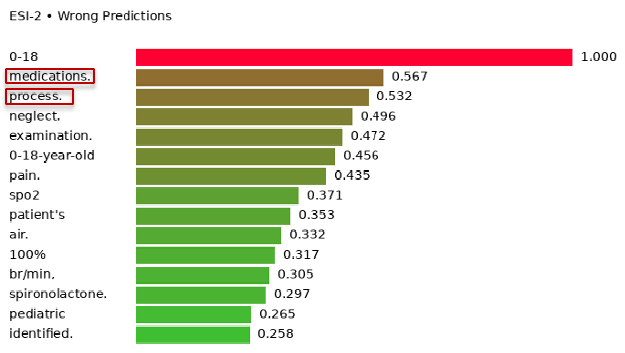}
\end{minipage}
\begin{minipage}{0.24\textwidth}
\centering
\includegraphics[width=\linewidth]{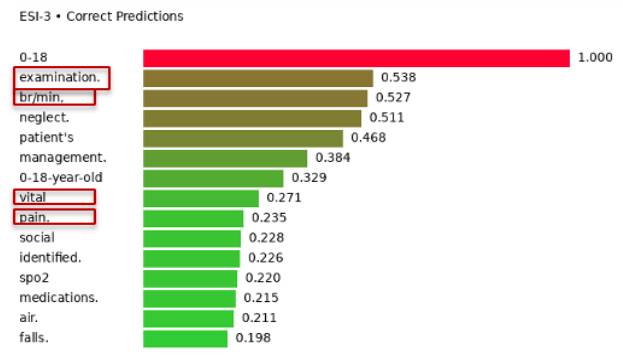}
\end{minipage}
\begin{minipage}{0.24\textwidth}
\centering
\includegraphics[width=\linewidth]{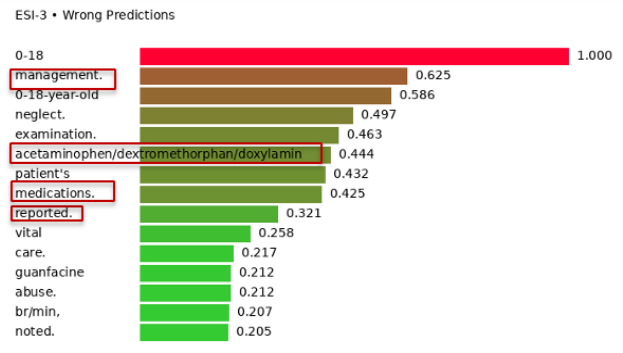}
\end{minipage}
\caption{Token-level attention patterns for correct vs. incorrect ESI predictions across ESI-2 and ESI-3.}
\label{fig:attention_patterns}
\end{figure}

To interpret model behavior, we analyzed token-level attention distributions for correctly and incorrectly classified encounters across ESI-2 and ESI-3 (Figure~\ref{fig:attention_patterns}). Several consistent patterns emerged.

\textbf{Correct predictions showed strong attention to clinically meaningful features} such as respiratory descriptors (e.g., ``br/min,'' ``vital,'' ``examination''), pain-related symptoms, and age-specific context (e.g., ``18-year-old,'' ``10-year-old''). These terms correspond to objective markers emphasized in ESI guidelines, including respiratory distress, abnormal vital signs, and concerning physical examination findings.

In contrast, \textbf{misclassified cases placed greater attention on nonspecific contextual or administrative cues}, such as ``management,'' ``process,'' ``reported,'' ``home,'' and medications (e.g., ``spironolactone,'' ``acetaminophen / dextromethorphan''). These tokens describe background history or care processes rather than indicators of acuity, suggesting that the model may occasionally be distracted by narrative details unrelated to triage severity.

Across ESI levels, incorrect predictions also showed increased attention to broad descriptors (e.g., ``child,'' ``pediatric,'' ``social''), vital-sign formatting tokens (e.g., ``97\%,'' ``(97.7$^\circ$F)''), or repeated structural patterns (e.g., ``0--18,'' ``identified''). This pattern indicates that the model can sometimes anchor on superficial lexical cues rather than clinically meaningful signals.

\section{Discussions}

\subsection{Non-LLM Baselines}
We first evaluated non-LLM baseline models for ESI prediction from clinical vignettes, including CNN, bidirectional LSTM RNN, BERT, fine-tuned BERT, ClinicalBERT, fine-tuned ClinicalBERT, and Random Forest. As shown in Table~\ref{tab:clinical_vignette_all}, among the non-LLM baselines, fine-tuned BERT achieved the lowest total discordance at 29.26\%, while fine-tuned ClinicalBERT achieved the lowest over-triage rate among neural transformer-based baselines at 14.77\%. Random Forest achieved the lowest over-triage and significant over-triage rates overall, but with higher under-triage and significant under-triage. These results suggest that while task-specific fine-tuning improves traditional neural and transformer-based baselines, domain-adapted small language models provide stronger overall performance for clinical vignette-based ESI prediction.

\begin{table}[t]
\centering
\scriptsize
\setlength{\tabcolsep}{2pt}
\caption{Comparison of non-LLM baselines, LLMs, fine-tuning, and RAG strategies for ESI prediction from clinical vignettes. Lower values indicate fewer triage errors. Runtime is reported in seconds per patient encounter when available.}
\label{tab:clinical_vignette_all}
\renewcommand{\arraystretch}{1.12}

\resizebox{\columnwidth}{!}{%
\begin{tabular}{l l r r r r r r}
\toprule
\textbf{Setting} & \textbf{Method} & \textbf{Disc.} & \textbf{Under} & \textbf{Over} & \textbf{Sig. Under} & \textbf{Sig. Over} & \textbf{Time} \\
\midrule

\multirow{12}{*}{\begin{tabular}[c]{@{}l@{}}clinical vignette\end{tabular}}
& \multicolumn{7}{l}{\textit{Non-LLM baselines}} \\
\cline{2-8}
& CNN & 38.64\% & 19.89\% & 18.75\% & 5.68\% & 8.24\% & 0.001 \\
\cline{2-8}
& Bidirectional LSTM & 30.40\% & 14.20\% & 16.19\% & 5.11\% & 7.10\% & 0.003 \\
\cline{2-8}
& BERT & 59.09\% & 22.73\% & 36.36\% & 5.11\% & 24.72\% & 0.001 \\
\cline{2-8}
& Fine-tuned BERT & 29.26\% & 12.22\% & 17.05\% & 4.26\% & 7.10\% & 0.006 \\
\cline{2-8}
& ClinicalBERT & 55.11\% & 21.02\% & 34.09\% & 5.97\% & 18.18\% & 0.0005 \\
\cline{2-8}
& Fine-tuned ClinicalBERT & 30.40\% & 15.62\% & 14.77\% & 5.97\% & 3.69\% & 0.004 \\
\cline{2-8}
& Random Forest & 32.95\% & 23.86\% & 9.09\% & 10.51\% & 0.28\% & 0.003 \\
\cline{2-8}

& \multicolumn{7}{l}{\textit{LLM, fine-tuning, and RAG strategies}} \\
\cline{2-8}
& GPT-4o & 56.98\% & 7.41\% & 49.57\% & 2.28\% & 27.64\% & 6.15 \\
\cline{2-8}
& GPT-4o + RAG & 48.15\% & 10.26\% & 37.89\% & 3.99\% & 13.96\% & 8.40 \\
\cline{2-8}
& Qwen2.5-7B & 51.70\% & 20.74\% & 30.97\% & 11.36\% & 6.25\% & 0.32 \\
\cline{2-8}
& \textbf{Fine-tuned Qwen2.5-7B} & \textbf{25.85\%} & 13.35\% & \textbf{12.50\%} & 6.25\% & \textbf{2.56\%} & \textbf{0.16} \\
\cline{2-8}
& Qwen2.5-7B + RAG & 68.47\% & 15.06\% & 53.41\% & 8.24\% & 32.95\% & 7.02 \\
\cline{2-8}
& Fine-tuned Qwen2.5-7B + RAG & 66.19\% & \textbf{5.40\%} & 60.80\% & \textbf{4.26\%} & 40.34\% & 12.59 \\
\bottomrule
\end{tabular}%
}
\vspace{-0.5em}
\end{table}

\begin{table}[t]
\centering
\scriptsize
\setlength{\tabcolsep}{2pt}
\caption{
Performance of one-round and two-round multi-agent configurations applied to the best fine-tuned Qwen2.5-7B model for ESI prediction from clinical vignettes. 
Multi-agent ensembles (3–4 agents) incorporate specialized clinical personas and majority voting, but—as shown—do not outperform the single-agent model. 
Lower values indicate better error rates.
}
\label{tab:Multi-Agent}
\renewcommand{\arraystretch}{1.12}

\resizebox{\columnwidth}{!}{%
\begin{tabular}{l l r r r r r r}
\toprule
\textbf{Setting} & \textbf{Method} & \textbf{Disc.} & \textbf{Under} & \textbf{Over} & \textbf{Sig. Under} & \textbf{Sig. Over} & \textbf{Time} \\
\midrule

\multicolumn{8}{l}{\textbf{Baseline (Single-Agent)}} \\
\midrule
Clinical vignette 
& Fine-tuned Qwen2.5-7B 
& \textbf{25.85\%} & \textbf{13.35\%} & \textbf{12.50\%} 
& \textbf{6.25\%} & \textbf{2.56\%} & \textbf{0.15} \\

\midrule
\multicolumn{8}{l}{\textbf{Multi-Agent (One-Round Voting)}} \\
\midrule

Clinical vignette
& Qwen2.5-7B + 3-Agent (1 round) & 33.24\% & 25.00\% & 8.24\% & 10.23\% & 1.70\% &  49.66\\
& Qwen2.5-7B + 4-Agent (1 round) & 33.24\% & 25.00\% & 8.24\% & 10.23\% & 1.70\% & 56.22 \\

\midrule
\multicolumn{8}{l}{\textbf{Multi-Agent (Two-Round Debate + Voting)}} \\
\midrule

Clinical vignette
& Qwen2.5-7B + 3-Agent (2 rounds) & 43.75\% & 16.76\% & 26.99\% & 6.53\% & 13.64\% & 21.42 \\
& Qwen2.5-7B + 4-Agent (2 rounds) & 43.47\% & 16.76\% & 26.70\% & 6.53\% & 13.64\% & 28.32 \\

\bottomrule
\end{tabular}%
}

\vspace{0.2em}
\begin{minipage}{\columnwidth}
\scriptsize
\textit{Notes.}  
Multi-agent systems combine specialized Qwen2.5-7B agents via majority voting (ties → safer ESI). 
Two-round variants allow agents to revise decisions after sharing rationales.
\end{minipage}

\end{table}

\subsection{Retrieval-Augmented Generation Exploration.}
To assess whether retrieval-augmented generation (RAG) improves ESI prediction, we compared four strategies using Qwen2.5-7B: (1) direct prompting, (2) fine-tuned prompting, (3) RAG with the \textit{ESI Handbook}, and (4) RAG after fine-tuning. As shown in Table~\ref{tab:clinical_vignette_all}, \textbf{fine-tuning alone provided the clearest benefit}, substantially reducing total discordance, over-triage, and significant over-triage errors, indicating effective alignment with clinical decision patterns. In contrast, \textbf{adding RAG increased over-triage and significant over-triage} even when combined with the fine-tuned model, and substantially increased runtime. These findings suggest that for structured, rule-based tasks such as ESI classification, naïve retrieval may introduce uncertainty rather than improve accuracy. For small language models, \textbf{high-quality fine-tuning may be more impactful than external knowledge retrieval}. By comparison, GPT-4o showed modest benefit from retrieval: supplementing it with the \textit{ESI Handbook} reduced discordance from \textbf{56.98\% to 48.15\%} (Table~\ref{tab:clinical_vignette_all}), mainly through fewer over-triage errors, though with slight increases in under-triage and significant under-triage. This suggests that the value of retrieval is \textbf{model-specific}. Future work will explore more targeted retrieval strategies (e.g., relevance-filtered or triage-aware retrieval) to reduce noise and improve reliability.

\subsection{Multi-Agent Exploration}
We evaluated multi-agent ensembles built on the best fine-tuned Qwen2.5-7B model using up to four specialized agents reflecting different triage perspectives: \textit{safety-first} (minimizes under-triage), \textit{guideline-strict} (follows ESI rules), \textit{resource-aware} (estimates resource needs), and \textit{red-flag/vitals sentinel} (detects abnormal vitals and life-threatening indicators). Ensembles were nested: the 3-agent system used safety-first, guideline-strict, and resource-aware; the 4-agent system used all agents.

In the one-round setting, agents independently predicted an ESI level with majority voting (ties resolved to the safer lower ESI). A two-round variant allowed agents to revise predictions after viewing others’ rationales. As shown in Table~\ref{tab:Multi-Agent}, multi-agent systems did not outperform the single-agent model. One-round ensembles increased discordance (~33\%) and under-triage (25\%), while two-round debate performed worse (discordance $>$43\%, significant over-triage 14–16\%). Runtime also increased substantially (21–56s vs. 0.15s). Overall, agent diversity introduced variability that degraded ESI accuracy; future work may explore structured coordination (e.g., role-weighted voting or guideline-constrained reasoning).

\subsection{Incomplete information} 
Table~\ref{tab:ablation_missing_info} evaluates model robustness when key clinical information is removed from the vignette. GPT-4o showed slightly lower discordance when physical exam or vital signs were removed (56.98\% to 48.4\%), but with shifts in error distribution, including increased undertriage without the physical exam. Direct prompting with Qwen2.5-7B remained relatively stable but maintained higher discordance ($\sim$52–54\%). In contrast, the fine-tuned Qwen2.5-7B model achieved substantially lower discordance (25.85\%) and only moderate degradation when physical exam or vital signs were removed (28.98\% and 29.83\%). These results highlight the potential of domain-adapted small language models to maintain reliable triage predictions even under incomplete clinical documentation, a common scenario in real-world emergency department settings.

\begin{table}[t]
\centering
\scriptsize
\setlength{\tabcolsep}{2pt}
\caption{Impact of missing clinical information on ESI prediction performance. Results compare GPT-4o, direct prompting, and fine-tuned Qwen2.5-7B when physical exam or vital signs are removed from the clinical vignette. Lower values indicate fewer triage errors.}
\label{tab:ablation_missing_info}
\renewcommand{\arraystretch}{1.12}

\resizebox{\columnwidth}{!}{%
\begin{tabular}{l l r r r r r r}
\toprule
\textbf{Model} & \textbf{Condition} & \textbf{Disc.} & \textbf{Under} & \textbf{Over} & \textbf{Sig. Under} & \textbf{Sig. Over} & \textbf{Time} \\
\midrule

\multirow{3}{*}{GPT-4o} 
& Complete & 56.98\% & 7.41\% & 49.57\% & 2.28\% & 27.64\% & 6.15\\
& (-) Physical Exam & 51.0\% & 13.1\% & 37.9\% & 10.3\% & 22.8\% & 12.4 \\
& (-) Vital Signs & 48.4\% & 11.4\% & 37.0\% & 8.3\% & 17.4\% & 14.5 \\

\midrule

\multirow{3}{*}{Qwen2.5-7B (Prompt)}
& Complete & 51.70\% & 20.74\% & 30.97\% & 11.36\% & 6.25\% & 0.32 \\
& (-) Physical Exam & 53.41\% & 21.31\% & 32.10\% & 13.07\% & 5.11\% & 7.30 \\
& (-) Vital Signs & 53.98\% & 23.86\% & 30.11\% & 13.07\% & 4.26\% & 8.95 \\

\midrule

\multirow{3}{*}{Qwen2.5-7B (Fine-tuned)}
& Complete & \textbf{25.85\%} & \textbf{13.35\%} & \textbf{12.50\%} & \textbf{6.25\%} & \textbf{2.56\%} & \textbf{0.16} \\
& (-) Physical Exam & 28.98\% & 14.49\% & 14.49\% & 6.53\% & 4.26\% & 8.90 \\
& (-) Vital Signs & 29.83\% & 15.91\% & 13.92\% & 6.82\% & 4.55\% & 8.78 \\

\bottomrule
\end{tabular}%
}
\vspace{-0.5em}
\end{table}

\section{Conclusion and Future Work}
Our results demonstrate that open-source small language models, when carefully fine-tuned on institution-specific pediatric triage data, can substantially improve the accuracy, stability, and safety of automated ESI assignment. Qwen2.5-7B offered the most favorable combination of performance and efficiency, outperforming larger models, naïve retrieval, and multi-agent variants while maintaining sub-second inference times suitable for real-world clinical workflows. These findings suggest that lightweight, privacy-preserving, and reproducible SLMs represent a practical pathway toward scalable triage decision support. Future research should extend evaluation to noisier real-time triage notes, include currently excluded patient populations, incorporate outcome-anchored ground truth, and assess clinician-in-the-loop performance to characterize better the operational impact of the SLM-based triage systems.

We acknowledge that this study has several limitations that could be further addressed in future studies. First, it relied on retrospective single-center pediatric ED data, which may limit generalizability to other institutions, patient populations, or triage workflows. Second, evaluation required encounters with relatively complete documentation (chief complaint, vital signs, and physical exam), whereas real-world triage notes are often sparse; therefore, reported performance likely represents an upper-bound estimate under cleaner documentation conditions. Third, primary mental and behavioral health presentations and ESI~1 encounters were excluded due to alternative triage pathways and nonstandard documentation, reducing applicability to these patient groups.
A fourth limitation is the use of nurse-assigned ESI levels as the reference standard. Real-world triage demonstrates substantial variability and may deviate from guideline-based care; for example, ESI~3 overtriage rates at the study site routinely exceed 25\%. Models trained on these labels may therefore reproduce existing practice patterns, making it difficult to distinguish model error from variability in human triage decisions. Future work from our group aims to address this limitation through the development of an outcome-anchored triage benchmark dataset that evaluates triage accuracy using objective clinical outcomes and resource utilization rather than nurse-assigned labels alone.  As an additional ongoing quality check, pediatric emergency physician reviewers are independently reviewing the evaluation cases to compare physician-adjudicated ESI with nurse-assigned ESI. In an initial review of 100 cases, physician reviewers agreed with the nurse-assigned ESI in 74\% of cases. Once completed for all 353 encounters, this physician-adjudicated standard will provide an additional benchmark for evaluating model performance beyond nurse-assigned ESI alone. Finally, this study did not evaluate prospective deployment or downstream clinical impact.

\makeatletter
\renewcommand{\@biblabel}[1]{\hfill #1.}
\makeatother

\bibliographystyle{vancouver}
\bibliography{conference_101719}  

\end{document}